\documentclass{article}

\usepackage{arxiv}

\usepackage[utf8]{inputenc} 
\usepackage[T1]{fontenc}    
\usepackage{hyperref}       
\usepackage{url}            
\usepackage{booktabs}       
\usepackage{amsfonts}       
\usepackage{nicefrac}       
\usepackage{microtype}      
\usepackage{lipsum}

\usepackage{graphicx}
\usepackage{csquotes}
\usepackage{rotating}
\usepackage{pifont}
\usepackage{tabularx}
\usepackage{booktabs}

\title{A Review of Visual Odometry Methods and Its Applications for Autonomous Driving}

\author{
  Kai Li~Lim\\
  The REV Project\\
  The University of Western Australia\\
  Perth, Australia\\
  \texttt{kaili.lim@uwa.edu.au} \\
  \And
  Thomas~Bräunl\\
  The REV Project\\
  The University of Western Australia\\
  Perth, Australia\\
  \texttt{thomas.braunl@uwa.edu.au} \\
}

\begin{document}
\maketitle

\begin{abstract}
The research into autonomous driving applications has observed an increase in computer vision-based approaches in recent years. In attempts to develop exclusive vision-based systems, visual odometry is often considered as a key element to achieve motion estimation and self-localisation, in place of wheel odometry or inertial measurements. This paper presents a recent review to methods that are pertinent to visual odometry with an emphasis on autonomous driving. This review covers visual odometry in their monocular, stereoscopic and visual-inertial form, individually presenting them with analyses related to their applications. Discussions are drawn to outline the problems faced in the current state of research, and to summarise the works reviewed. This paper concludes with future work suggestions to aid prospective developments in visual odometry.
\end{abstract}

\keywords{Visual odometry \and Computer vision \and Autonomous Driving}

\section{Introduction}
Autonomous driving has come a long way since it was first promoted in the DARPA Urban Challenge back in 2007. With major car manufacturers lobbying their technologies in autonomous driving, the ownership of autonomous vehicles is set to rise in the future. Current autonomous vehicles rely on a variety of sensors to achieve self-localisation and obstacle avoidance. These can include a combination of laser scanners, radar, GPS, and camera. However, the installation of sensor arrays on a vehicle greatly increases its cost and complexity. At the same time, the increasing affordability and ubiquity of cameras and high-performance graphics processing units (GPUs) are catalysing the resurgence of image processing and computer vision applications. In other words, these applications that were once computationally expensive, are gradually replacing tasks that were performed using other sensors and methods. These tasks include the motion estimation of the vehicle, where precise odometry is crucial for the accurate localisation of the autonomous vehicle. The odometry problem exists such that conventional GPS sensors are unable to provide the necessary road lane precision (\textasciitilde3 m), and that it is unable to function indoors such as inside tunnels and buildings. Additionally, standard wheel odometry suffers from accumulating drift errors that increase over time. While the use of sensors such as high precision differential GPS and inertial sensors could alleviate this problem, they are significantly more expensive to purchase than a standard camera setup. 

Visual odometry (VO) is a research area that is becoming increasingly popular in recent years. Ground vehicles and robots rely on odometry to measure and record their traversed path as they navigate, making this an essential component for autonomous navigation. Visual odometry is odometry that is performed by analysing visual data such as one from a mounted camera. This concept was first proposed by Moravec in \cite{RN103}, and the term ``visual odometry" was coined by Nistér et al. in \cite{RN20}. Conventional wheel odometry estimates a robot's position by measuring the wheel rotation using sensors from the servos. A common issue experienced by wheel odometry is wheel slip, whereby pose estimations becomes incrementally inaccurate from the occasional loss of traction from the wheels. Visual odometry negates this problem.

VO techniques can be classified according to their utilised imagery --- either stereoscopic or monocular visual odometry, and their processing techniques --- either feature-based or direct (image/appearance-based). VO methods can either use a combination of feature matching, feature tracking or optical flow  \cite{RN22, RN25}. Since a majority of visual odometry implementation recreates a 3D navigation environment from a set of captured images, most approaches are of a stereoscopic approach that utilises a pair of mounted cameras on the robot. By accounting for the cameras' capture frame rate and the distance between them, the robot's displacement and velocity from an object can be calculated with ease through the triangulation of image features \cite{hartley_triangulation_1997}. Therefore, the monocular visual odometry problem is more complex and it is not until recently that we are starting to see an increasing trend in this area. Monocular visual odometry achieves motion estimation and environment recreation through a combination of a series of at least three 2D images in series, along with its bearing data. An adaptation of the parallel tracking and mapping (PTAM) algorithm \cite{RN98} is used in many monocular implementations. PTAM is originally devised for augmented reality (AR) implementations, but its speed and robustness while relying only on existing map features made it a popular choice for researchers of visual odometry. 

On the processing techniques front, feature-based approaches achieve motion estimation by extracting image features such as lines and edges, and tracking them in subsequent frames; by calculating the Euclidean distances of each feature points between frames, the displacement and velocity vectors can be calculated. Direct approaches use pixels in an image frame and track the changes in pixel intensity \cite{valiente_garcia_visual_2012}, where pixel selection can either be all pixels (dense) or sparsely selected (sparse). In feature-based approaches, feature matching detects and tags existing features on a given set of frames. Feature extraction and matching techniques such as Scale Invariant Feature Transform (SIFT) \cite{RN100}, Features From Accelerated Segment Test (FAST) \cite{RN99}, Speeded Up Robust Features (SURF) \cite{RN102}, Binary Robust Independent Elementary Features (BRIEF) \cite{RN104} and Oriented FAST and Rotated BRIEF (ORB) \cite{RN101} are some of the more commonly implemented ones in literature. Feature tracking techniques allow features to be tracked across subsequent frames. This is usually used in tandem with features obtained from a feature extraction technique. Feature tracking is essential for visual odometry, as it allows the robot to achieve a consistent measurement to localise itself \cite{RN25}. Varying conditions in the environment such as lighting conditions and dynamic obstacles can impede accuracy with outliers. To circumvent this, many works employ the Random Sampling Consensus (RANSAC) \cite{RN105} outlier rejection scheme or a variation of it; more recently, Buczko and Willert have also proposed an outlier detection scheme for monocular \cite{buczko_monocular_2017} and stereoscopic \cite{buczko_how_2016} approaches. Finally, optical flow allows the robot to estimate its distance from an environmental object by tracking its features from the robot's camera feed. With optical flow, the robot can perform obstacle detection and avoidance during navigation. An optical flow algorithm outputs an image pattern that relates to the movement of objects within the robot's field of vision (FOV). Examples of popular optical flow algorithms include the Lucas and Kanade's \cite{RN108}, Horn and Schunck's \cite{RN110}, Farneback's \cite{RN111}, and SimpleFlow \cite{RN36} algorithms. Optical flow algorithms can either be dense (tracks a full frame) or sparse (tracks extracted features). Dense optical flow requires greater computation performance, whereas sparse optical flow methods employ feature extraction prior to its computation to make it less intensive. 

It is also common for researchers to use a combination of the three processing techniques to achieve robust optical flow. For example, Wang and Schmid \cite{RN134} used a combination of SURF descriptors, RANSAC for outlier rejection, and dense optical flow to achieve the prediction of human actions. Liu et al. \cite{RN135} proposed an optical flow approach based on the Maximum Likelihood Estimation (MLE), which is implemented on a mobile robot and compared against their RANSAC development for optical flow; they concluded that their MLE approach is more accurate than the RANSAC approach. More recently, Kroeger et al. \cite{RN27} proposed a faster approach for dense optical flow computation using the dense inverse search (DIS) \cite {RN136} method, noting that many optical flow proposals have neglected time complexity in favour of accuracy. The authors' evaluation of the DIS fast optical flow showed that while it introduced slight estimation errors, it is much faster even when compared to newer optical flow methods. 

This paper reviews monocular and stereoscopic VO methods according to their procedures to achieve motion estimation, as well as their methods of evaluation. A shorter section on visual-inertial odometry is also presented to explore works that combine inertial measurements for VO. The review of these VO methods is intended to gauge their suitability for use in real-time, on-line autonomous driving, which is motivated by the research gap in VO applications for autonomous vehicles. The aim of this article is hence to understand the current trends in VO and to determine if the current state of VO is adequate enough to be utilised in autonomous vehicles. 

\begin{figure}[ht]
	\centering
	\includegraphics[width=0.5\linewidth]{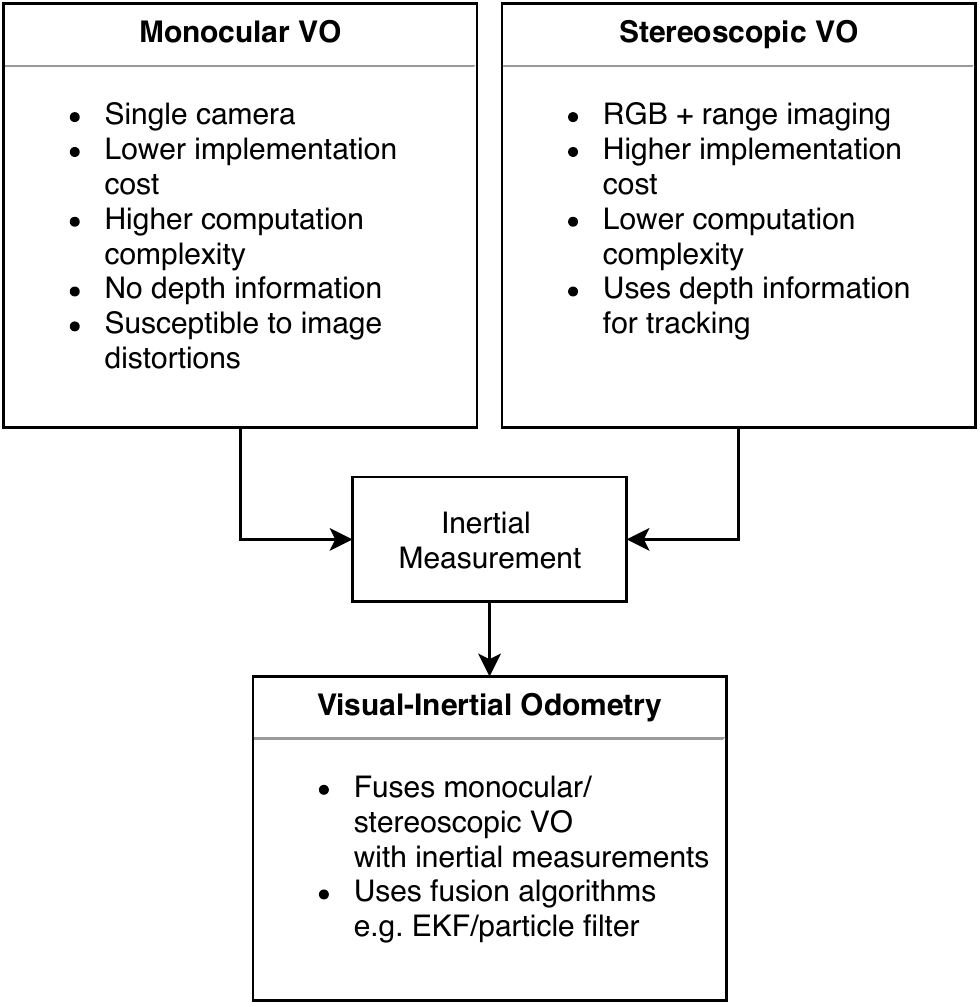}
	\caption{\label{figimagery}Types of VO approaches as classified by their imagery methods. Visual-inertial odometry is VO that is fused with inertial measurements.}
\end{figure}

\begin{figure}[ht]
	\centering
	\includegraphics[width=0.9\linewidth]{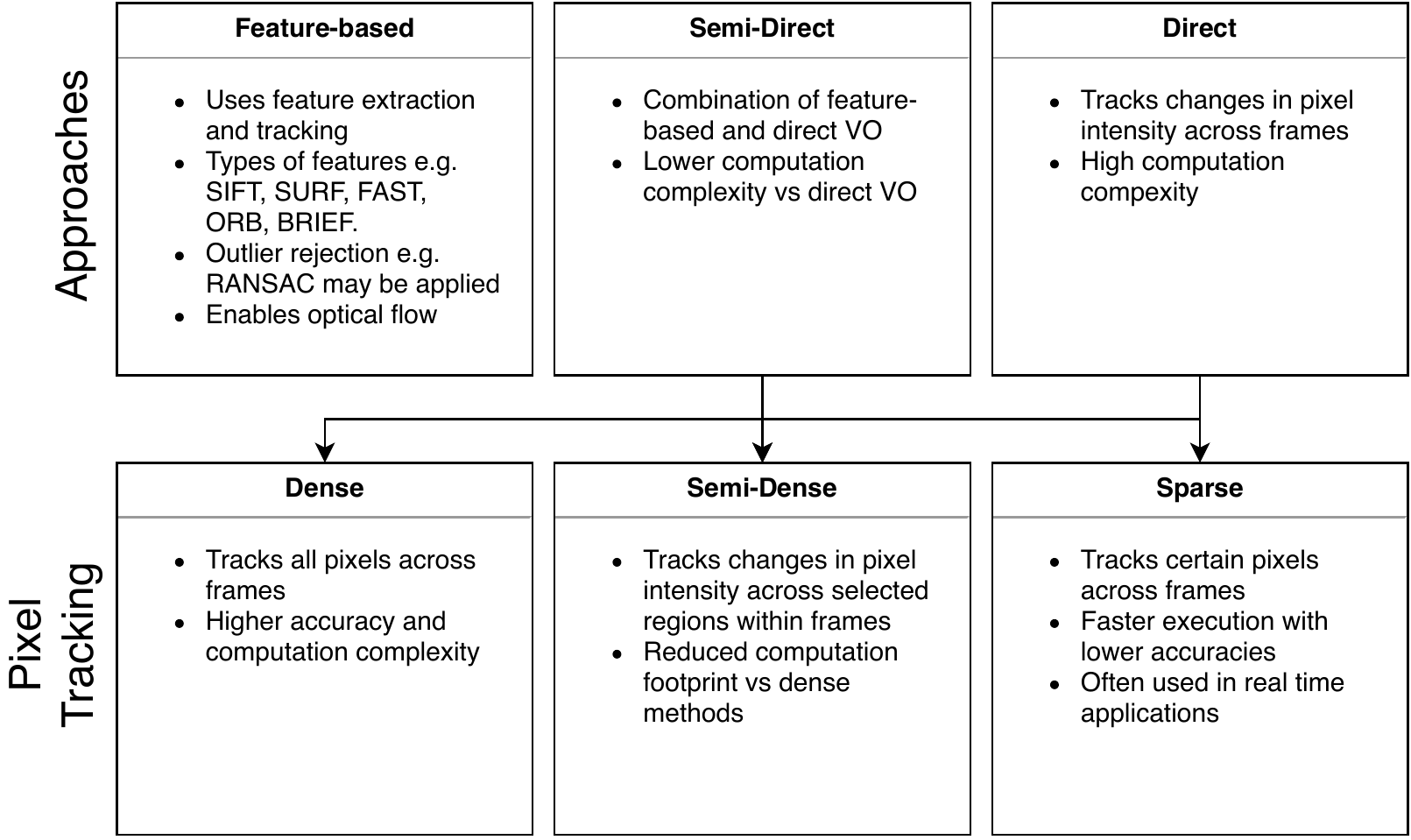}
	\caption{\label{figprocessing}Processing techniques used in VO according to their approaches and their pixel tracking methods for direct approaches. }
\end{figure}

Figures~\ref{figimagery} and ~\ref{figprocessing} summarise the scope of this paper. The remainder of this paper is organised according to imagery techniques given in  Figure~\ref{figimagery}, whereby Sections~\ref{secmvo} through~\ref{secvio} describes monocular VO, stereoscopic VO and visual-inertial odometry respectively; Section~\ref{secdiscuss} presents critical discussions from our review findings in relation to autonomous driving before the conclusion is drawn in Section~\ref{secconcl}. Reviews will emphasise on the VO approaches and for direct methods, the pixel tracking density according to Figure~\ref{figprocessing}, which is further elaborated in Section~\ref{secmvo}.

\section{Monocular Visual Odometry}\label{secmvo}
Using a monocular camera setup for VO benefit implementations that are lower in cost and complexity. A monocular setup will also alleviate the decrease in depth measurement accuracies as the distance between the camera and the scene increases beyond the stereo baseline. This setup, however, introduces several challenges in addition to the lack of depth measurements on a stereoscopic setup. This was pointed out by Yang et al. \cite{yang_challenges_2017}, where they have investigated several challenges that pertain to this area namely photometric calibration, motion bias, and (assuming that a roller shutter camera is used) the rolling shutter effect. Photometric calibration is required as the pixel intensity for a same 3D point will experience varying values due to the changes in camera adjustment such as optical exposures and gains; motion bias notes that VO performances are different for forward and backward playback on the same sequence; and the rolling shutter effect is predominantly present in rolling shutter cameras whereby an image will distort in while a camera is in motion as the frame is captured line by line. The authors then analysed these challenges using a feature-based method, a semi-direct method, and a direct method, which are ORB-SLAM \cite{mur-artal_orb-slam2:_2017}, SVO \cite{forster_svo:_2017} and DSO \cite{engel_direct_2018} respectively. The several conclusions drawn from this analysis allowed us to deduce that direct methods are more robust with photometric calibration while being insensitive to pixel discretisation artefacts; while it is affected by the rolling shutter effect, in terms of autonomous driving, using a global shutter camera will nullify this. 

For feature-based VO, Chien et al. \cite{chien_when_2016} compared the SIFT, SURF, ORB and A-KAZE \cite{alcantarilla_fast_2013} feature extraction methods for monocular VO. Experiments were conducted on the KITTI dataset using OpenCV 3.1, and concluded that while SIFT is the most accurate at extracting features, ORB is less computationally intensive, the A-KAZE method sits between SIFT and ORB in computational requirements and accuracy. We hence decided that as our autonomous driving implementation uses an embedded computer, the ORB method is better suited for our applications. 

Prominent monocular VO algorithms that are recently proposed include Direct Sparse Odometry (DSO) \cite{engel_direct_2018} and Semidirect Visual Odometry (SVO) \cite{forster_svo:_2017}. As their names suggest, DSO uses a direct approach whereas SVO uses a semi-direct approach to monocular VO. DSO also uses a sparse formulation thereby decreasing computation complexity, as opposed to dense \cite{newcombe_dtam:_2011, meilland_real-time_2011} and semi-dense \cite{daftry_semi-dense_2015, engel_lsd-slam:_2014} formulations of past proposals. This meant that DSO is capable of achieving real-time computation, as it samples only points of sufficient intensity gradient, and neglecting the geometric prior. DSO functions by continuously optimising photometric parameters from the camera to achieve photometric calibration. This optimisation was performed using a Gauss-Newton method through a sliding window. DSO uses sparse technique whereby it samples data points that are of a limited and equally distributed number across space and active frames, thereby reducing sampling redundancy for data point management. Experimental results showed that DSO as a direct approach is robust against photometric noise, and is able to achieve high accuracies with proper calibration. 

SVO was proposed to solve the slow computations and lack of optimality and consistencies of direct methods by combining traits of direct and feature-based methods. This algorithm performs a minimisation of photometric errors on features of the same 3D point, where subpixel features are subsequently obtained through the relaxation of geometric constraints. The minimisation of photometric error is performed at the sparse image alignment stage using a method of least squares, where it assumes that depth information is only known at corners and features that lie on intensity edges. A sparse method uses little depth information and hence the authors enhanced its robustness by aggregating the photometric cost for pixels surrounding the feature, with approximations similar to the feature depth. SVO employs drift minimisation by relaxing geometric constraints and aligning corresponding feature patches to an older reference patch, which is subsequently optimised for reprojection errors using a bundle adjustment. To improve computation efficiency, SVO uses a second thread for mapping, which initialises a new depth filter at FAST corners at every keyframe, thereby estimating the pixel depth using a recursive Bayesian depth filter. Experiments comparing SVO against ORB-SLAM and LSD-SLAM showed that SVO is more efficient at tracking features due to its sparse approach while being robust to high-speed camera captures without the need for outlier rejection methods such as RANSAC. For application requiring high accuracies, the authors used iSAM2 \cite{kaess_isam2:_2012}, which applies incremental smoothing for the trajectory motion, thereby achieving the same accuracy as a batch estimation of the entire trajectory in real-time. SVO can also be fused with inertial measurement to further increase odometric accuracies. 

\subsection{Related Applications}
Monocular VO algorithms are often tested as a benchmark on datasets such as MonoVO \cite{engel_photometrically_2016} and KITTI's monocular VO dataset \cite{a._geiger_are_2012}. Recent works that implement monocular VO include the work by Sappa et al. \cite{sappa_monocular_2016}, which uses fused images to achieve monocular VO through a discrete wave transform (DWT) scheme where the characteristics of the captured image determine the DWT parameters. Using an image fusion technique condenses information from multiple image frames into one before it is used for motion estimation. This method is compared against VISO2 \cite{kitt_visual_2010}, and experiments were performed on video sequences captured on vehicles driving at different times of day and location. By comparing this fusion approach with previous approaches, the authors noted that the algorithm performs well in challenging environments such as low light drives. Results during daytime are similar across all compared approaches.  

The online supervised approach presented by Lee et al. \cite{lee_online_2015} uses ground classification to achieve monocular VO. The authors employed an appearance-based approach with RANSAC over three successive frames to obtain the image flow. Online self-learning is achieved by combining geometric estimates with the ground classifier, which uses a histogram of colour labels. The authors tested their approach on the KITTI odometry dataset where it is compared against the VISO2 algorithm and concluded that their approach was superior in terms of stability and translation performance. While the exclusive use of ground information is adequate to achieve VO for autonomous driving, the authors noted that it could be worthwhile to extend the work to estimate other object models. 

Additionally, works that combine semantic segmentation and monocular VO include \cite{ros_vision-based_2015, an_semantic_2017}. An et al. \cite{an_semantic_2017} introduced a semantic-segmentation aided VO in order to identify and compensate for dynamic visual obstructions in a camera frame. A modified version of SegNet \cite{badrinarayanan_segnet:_2015} is used to find visual cues that represent regions of actual motion. The VO method is a semi-direct method whereby in the feature-based section, the authors employed a k-nearest neighbour method to match keypoints from prior frames with its transformation solved using a least-square minimisation method; in the direct (alignment-based) method, the authors used a semi-dense method whereby the framework only utilises regions with certain segmentation labels to ensure that planar objects that are not in motion are selected, which are road and pavement regions, and road markings. Tests were performed on the KITTI odometry dataset and the authors' Beijing Wuhan dataset. The VO approach was compared against the VISO \cite{geiger_stereoscan:_2011}, DSO and ORB-SLAM2 algorithms, and concluded that a semantic segmentation approach is able to compensate moving objects on the road, where the VISO, DSO and ORB-SLAM2 could not. 

The approach by \cite{an_semantic_2017} ties into our requirements for our autonomous driving platform, as we also employ semantic segmentation to achieve visual autonomous driving. Also, the authors noted that ORB-SLAM2 achieved the best accuracy on the KITTI dataset on low traffic segments. This could also indicate that an ORB-SLAM-based method with semantic segmentation could be implemented for autonomous driving if it's properly optimised. If performance cost is an issue, we hypothesise that using a ground-only approach such as \cite{lee_online_2015} could achieve adequate VO for autonomous driving, while neglecting dynamic road objects. A robust day-night implementation could benefit from the implementation of a fusion technique as in \cite{sappa_monocular_2016}, but its overhead performance cost needs to be taken into consideration. A summary of the monocular VO algorithms reviewed is given in Table \ref{tablemono}, which lists its approach type, descriptors/features, outlier rejection scheme, dataset evaluated, and intended environment. 

\begin{sidewaystable}
	\caption{Summary of monocular VO methods reviewed}\label{tablemono}
	\centering
	\footnotesize
	\begin{tabularx}{\textwidth}{X c c c c c}
		\toprule
		\bfseries Method & \bfseries Approach & \bfseries Descriptor & \bfseries Outlier Rejection & \bfseries Dataset & \bfseries Environment \\
		\midrule
		DSO \cite{engel_direct_2018} & Direct & Photometric & Threshold filter & monoVO, EuRoC, ICL-NUIM & Aerial drone \\
		SVO \cite{forster_svo:_2017} & Semi-direct & Photometric + FAST & Depth filter & TUM RGB-D, EuRoC, ICL-NUIM, Circle \cite{forster_svo:_2017} & Aerial drone \\
		Sappa et al. \cite{sappa_monocular_2016} & Feature & SURF (VISO2) + DWT & RANSAC & KAIST \cite{choi_kaist_2018}, CVC \cite{poujol_visible-thermal_2016} & Road vehicle \\
		Lee et al. \cite{lee_online_2015} & Direct & Geometric estimate & RANSAC & KITTI & Road vehicle \\
		An et al. \cite{an_semantic_2017} & Semi-direct & Photometric + SegNet & RANSAC & KITTI, Beijing Wuhan \cite{an_semantic_2017} & Road vehicle\\
		\bottomrule
	\end{tabularx}
	
	\caption{Summary of stereoscopic VO works reviewed}\label{tablestereo}
	\centering
	\footnotesize
	\begin{tabularx}{\textwidth}{X c c c c c}
		\toprule
		\bfseries Method & \bfseries Approach & \bfseries Descriptor & \bfseries Outlier Rejection & \bfseries Dataset & \bfseries Environment \\
		\midrule
		CV4X \cite{persson_robust_2015} & Feature & FAST + BRIEF & RANSAC & KITTI & Road vehicles\\
		Stereo DSO \cite{wang_stereo_2017} & Direct & Photometric & Threshold filter & KITTI, Cityscapes & Road vehicles \\
		Wu et al. \cite{wu_framework_2017} & Feature & KLT & RANSAC & KITTI & Road vehicles\\
		Proen\c{c}a and Gao \cite{proenca_probabilistic_2018} & Feature & SURF + LSD & Circle matching & TUM RGB-D, ICL-NUIM & Pedestrian (Indoor)\\
		Holzmann, Fraundorfer and Bischof \cite{holzmann_detailed_2016} & Direct & SAD & Cauchy Loss function & Rawseeds \cite{ceriani_rawseeds_2009}, KITTI, \cite{holzmann_detailed_2016} & Aerial, ground robots\\
		Jaimez et al. \cite{jaimez_fast_2017} & Direct & Photometric + Background Seg & \textit{None} & TUM RGB-D & N/A\\
		Liu et al. \cite{liu_robust_2017} & Feature & SURF (VISO2) & Circle matching & KITTI, New Tsukuba & Road vehicles\\
		Kunii, Kovacs and Hoshi \cite{kunii_mobile_2017} & Feature & FREAK, CenSurE & RANSAC & Real world (Izu Oshima) & Mobile robot\\
		Sun et al. \cite{sun_sequentially_2017} & Feature & U-SURF & RANSAC & Real world & Mobile robot \\
		Kim and Kim \cite{kim_effective_2016} & Direct & Photometric & t-distribution & TUM RGB-D, real world & Mobile robot \\
		\bottomrule
	\end{tabularx}
\end{sidewaystable}

\section{Stereoscopic Visual Odometry}\label{secsvo}
Visual odometry is stereoscopic when a depth measurement is obtainable from a range imaging/RGB-D camera, often through a stereoscopic camera which uses stereo triangulation to measure depth. Other range imaging cameras that are used for stereoscopic VO include structured-light cameras \cite{liu_improved_2017,huang_visual_2017} and time-of-flight cameras \cite{stefan_threedimensional_2009}. This depth measurement enables the distance between the vehicle and its surrounding objects to be perceived, thereby simplifying VO calculations. It is also possible to rely solely on this depth measurement to perform VO, as opposed to using an image-based (feature or appearance-based) approach; this is known as a depth-based approach \cite{fang_experimental_2015}. Applications for stereoscopic VO are popular, having a list that includes the Mars rover \cite{maimone_two_2007} and autonomous aerial drones \cite{strydom_visual_2014}. Its popularity today can be attributed to the leaderboard on the KITTI VO benchmark \cite{geiger_visual_nodate}, whereby a great proportion of the most accurate methods are stereoscopic. As opposed to a monocular approach, stereo VO requires proper calibration and synchronisation of the camera pair, as errors will directly affect VO performance \cite{kitt_visual_2010}

Stereoscopic VO can also be classified from a combination of feature- or appearance-based approaches. A review was presented by Fang and Zhang \cite{fang_experimental_2015} in 2014, which compares several stereoscopic VO methods according to their approaches. The authors tested these methods empirically on the TUM RGB-D dataset \cite{sturm_benchmark_2012}, along with the authors' dataset collected from an indoor environment. This dataset tests the algorithms in a variety of environments with a combination of fast motion, illumination invariances and limited features. Stereo VO algorithms were measured for their accuracies and computational performances. Results showed that depth-based algorithms such as Rangeflow \cite{jaimez_fast_2015} is robust in environments that lack features or illumination; an image-based algorithm is suitable for feature-rich environments with adequate illumination. 

It should be noted that the implementation simplicity of stereo VO, when compared to the added computation complexity of monocular VO, could imply that the research advancements made by monocular VO are more substantial than that of its stereo counterpart; this argument was pointed out by Persson et al. \cite{persson_robust_2015}, to which they have presented a stereo VO algorithm dubbed the CV4X that leverages on monocular VO techniques. The authors selected a feature-based method using FAST descriptors for corner extraction that is filtered with an Adaptive Non-Maxima Suppression (ANMS) filter, and tracking is performed on BRIEF descriptors. The camera's pose estimation was performed using RANSAC for the perspective-{\it n}-point (PNP) problem. Stereo triangulation errors are minimised through an iterative minimising function. CV4X was tested on the KITTI VO dataset and subsequently achieving first place on the benchmark leaderboard at its time of publication. The performance of this algorithm was optimised using OpenMP \cite{openmp_openmp_nodate} and CUDA \cite{nvidia_corporation_parallel_2016} for subtask parallelisation during its experiments. 

Monocular algorithms that are adapted into a stereo method also exist, with recent examples including the Stereo DSO \cite{wang_stereo_2017}, where the authors noted that both stereo approaches are complementary, and that multi-view stereo is able to negate the limitation in depth measurements that occur in direct stereoscopy. Multi-view stereoscopy captures stereo images using two or more images \cite{furukawa_multi-view_2015}, whereas direct stereoscopy achieves this using only an image pair. Direct stereo is used here for the initial depth estimation for multi-view stereo. The tracking of features is achieved using direct image alignment \cite{forster_svo:_2017} that is optimised using a Gauss-Newton method. Experiments were performed on the KITTI VO and Cityscapes \cite{cordts_cityscapes_2016} datasets to evaluate tracking and 3D reconstruction. Results showed that the Stereo DSO yields low translational and rotational errors when compared to LSD-VO \cite{engel_lsd-slam:_2014} and ORB-SLAM2 \cite{mur-artal_orb-slam2:_2017}, with denser and more accurate 3D reconstructions. 

The work by Wu et al. \cite{wu_framework_2017} is one that combines pruned Kanade-Lucas-Tomasi (KLT) tracking \cite{tomasi_detection_1991} and Gauss-Netwon optimisation with RANSAC to achieve fast feature-based stereo VO. The proposed pruning-based corner detector reduces redundancies while achieving robust corner detections for feature tracking. The KLT tracker was subsequently optimised to track these features through the addition of a pyramidal process. Evaluations on the KITTI dataset showed that while the proposed method was not able to best the state-of-the-art on the leaderboard, its fast estimation process is able to ensure that VO is performed with a smaller computation footprint. 

Methods that rely on time-of-flight or structured-light cameras such as the Kinect sensor captures depth maps that are prone to noises that affect the accuracy of depth measurements. Feature tracking on depth maps can be simplistic due to its reduced detail and colour as compared to an RGB image. For example, Proen\c{c}a and Gao \cite{proenca_probabilistic_2018} recently proposed a minimalistic environmental representation that combines points, line and planes to achieve VO. The authors noted that the limited field-of-view and the presence of noise on the depth map, which prompted them to propose a method that performs noise reduction using the missing depth measurement recovery technique with and depth uncertainty modelling. The extraction of these features is done using SURF for points, LSD \cite{gioi_lsd:_2010} for lines and the plane model is fit through a segmented point cloud. Experiments were performed on the TUM RGB-D dataset and the ICL-NUIM \cite{handa_benchmark_2014} dataset, as well as a dataset collected by the authors. Results showed that this method outperforms the point and line VO methods, as well as DVO and FOVIS \cite{huang_visual_2017}, this method is unable to outperform methods such as \cite{newcombe_kinectfusion:_2011,gutierrez-gomez_dense_2016}.

Holzmann, Fraundorfer and Bischof \cite{holzmann_detailed_2016} proposed a line-based direct stereo VO method using vertical lines. Limiting detections to vertical line enhances detection speed thereby making it suitable for real-time applications. Line verticality is determined using an IMU or a gravity-aligned camera. These lines are matched at every keyframe with direct pose estimation to achieve fast VO. The authors noted that the algorithm performs well in man-made environments, especially indoors with walls and fixtures, even in poorly textured environments; this approach is also adequate for urban driving scenes with  well-defined buildings, vehicles and road edges, as experiments on the KITTI VO dataset revealed that this method achieves results that are comparable to VISO2. However, the heavy presence of textures in outdoor drive scenes prevented the accuracy of this method to surpass that of VISO2. 

Another recent method to achieve fast VO is through geometric clustering. Using geometric clustering solves the same problem with dynamic scenes as listed in \cite{an_semantic_2017}, whereby objects in the frame that are in motion are capable of distorting VO accuracies. The approach presented by Jaimez et al. \cite{jaimez_fast_2017} applies k-means clustering on image points, in addition to segmenting image regions where static objects such as road regions for VO. The authors calculated VO through the minimisation of photometric and geometric residuals between the stereo image pairs, which is then estimated using a Cauchy M-estimator. VO performances were evaluated on the TUM RGB-D dataset and are compared against DIFODO \cite{jaimez_fast_2015}, DVO and SR-Flow \cite{quiroga_dense_2014}. Results showed that while this approach performs remarkably on dynamic scenes, it is unable to outperform the accuracies of the other algorithms on static scenes. Nevertheless, this method is capable of fast runtimes and real-time performances, which was not achieved by the other compared algorithms. 

In order to improve outlier rejection in dynamic environments, the work presented by Liu et al. \cite{liu_robust_2017} presents a stereo VO method that aims to improve accuracies using an improved outlier rejection method. The PASAC method is an improvement of RANSAC that achieves higher accuracies through a series of procedure. The authors noted the degradation of RANSAC's accuracy on frames with many outliers, as well as its uniformly generated hypothesis through the sampling of input data, thereby motivating the proposal of PASAC though an enhancement to its hypothesis generation for increased outlier rejection speed and accuracy. As a feature-based method, this method first extracts corner-like features from a stereo image pair, which are then matched using a sum of absolute differences method. Outliers from the image pair and its subsequent frames are then rejected through circle matching to identify mismatched features before the features are tracked according to its detection timestamp. Motion estimation is subsequently performed by an iterative solving of the non-linear least square optimisation problem, with PASAC as the outlier rejection model. This approach was tested on the KITTI VO and New Tsukuba \cite{martull_realistic_2012} datasets, outperforming RANSAC and PROSAC \cite{chum_matching_2005} in execution speed and accuracy. 

\subsection{Related Applications}
An implementation of stereo VO was first described by Nistér et al. in 2006 \cite{nister_visual_2006}, which estimates the ego-motion of a camera mounted on an autonomous ground vehicle. Since then, recent stereo VO methods are mostly applied in the robotics field, where they are often implemented onto mobile robots and aerial drones. While datasets from KITTI and TUM are often used to evaluate new stereo VO algorithms, real-world VO applications on-road vehicles are, to our knowledge, quite scarce. For example, while a recent thesis by Aladem \cite{aladem_robust_2017} described VO for autonomous driving, its evaluations of VO is limited to a ground robot and datasets; an attempt was made to collect local data from a car-mounted camera, but it resulted in unfavourable VO results. For these reasons, we will look into VO methods that are implemented onto mobile ground robots, as its application is most similar to that of a ground vehicle. 

The application by Kunii, Kovacs and Hoshi \cite{kunii_mobile_2017} uses a feature-based stereo VO method on a mobile robot through landmark tracking. The authors extracted FREAK \cite{alahi_freak:_2012} descriptors using CenSurE \cite{agrawal_censure:_2008} after comparing various extraction and descriptor methods for its environment, and stereo-matching is performed using a sum of absolute differences; RANSAC is used as the outlier rejection scheme. By using these VO parameters, the authors compared VO performance against GPS data and deduced errors of less than 5\% in both 2D and 3D space. The VO method is enhanced using template matching to improve computation footprint and accuracy, which returns the robot's position relative to the obstacles ahead. By comparing several methods against a laser scanner, the Zero Mean Normalised Cross Correlation (ZNCC) method was used as it returns the least amount of deviation from the laser scanner. Field experiments confirmed that the addition of template matching is able to result in accurate localisation. 

Likewise, the application by Sun et al. \cite{sun_sequentially_2017} implements stereo VO on a differentially-driven mobile robot. The VO method is feature-based whereby feature extraction is done using an upright SURF method, and stereo-matching is done using the perspective-three-point method; RANSAC is used as the outlier rejection scheme, which is also applied to the perspective-three-point method to filter erroneous landmarks. The authors proceeded to describe a fuzzy model for driving the robot that utilises stereo VO for trajectory stabilisation. Experimental results showed that the robot was able to achieve accurate trajectories based on the utilisation of VO alongside the Takagi-Sugeno fuzzy model \cite{takagi_fuzzy_1985}.

Noting that applying VO on background sections of a frame can result in better accuracies in dynamic environments, the application by Kim and Kim \cite{kim_effective_2016} implements a depth-based dense VO onto a differential drive robot. This background model-based dense-visual-odometry (BaMVO) algorithm isolates moving objects in the foreground by using a nonparametric model in \cite{elgammal_non-parametric_2000}, measuring the depth differences for objects in consecutive frames. VO on the background model is performed similarly to DVO \cite{steinbrucker_real-time_2011}, using a minimisation of the weighted sum of squares method. Outlier rejection of the background model is performed using the method in \cite{kerl_robust_2013}, which filters outliers over a t-distribution. The BaMVO was evaluated on the TUM RGB-D dataset before it was implemented for trajectories captured on the mobile robot in a dynamic indoor environment with pedestrians. The algorithm was compared against DVO and other state-of-the-art, which resulted in a the BaMVO being the most accurate, especially in dynamic environments where the other algorithms had erroneously calculated VO based on moving objects. 

By evaluating the aforementioned methods with regards to autonomous driving, the contributions made by \cite{kunii_mobile_2017} involve using a template matching method to enhance VO accuracies can be adapted for autonomous driving whereby it can be used to estimate the vehicle's distance relative to its surrounding obstacles. By complementing ZNCC with stereo VO techniques, a more robust VO solution can be applied for autonomous driving. The proposal in \cite{sun_sequentially_2017} which utilises stereo VO to ensure accurate motion trajectories can be similarly applied for autonomous driving to complement the path planning module of the autonomous car. This means that the adjustments in trajectories can be dynamically adjusted using results from VO to result in added robustness for path planning. Finally, the takeaway from Kim and Kim's \cite{kim_effective_2016} application suggests that adopting a VO algorithm for dynamic environment enables the vehicle to filter moving road objects such as other vehicles to restrict VO calculations on static, background models, thereby increasing VO robustness. A summary of the stereoscopic VO algorithms reviewed is given in Table \ref{tablestereo}, which lists its approach type, descriptors/features, outlier rejection scheme, dataset evaluated, and intended environment. 

\section{Visual-Inertial Odometry}\label{secvio}
Visual-inertial odometry (VIO) is a technique whereby a VO method is fused with the output from an inertial measurement unit (IMU) in order to improve odometry accuracy. Most VO algorithms can be adapted for VIO, resulting in VIO approaches that are either stereoscopic or monocular; using direct, semi-direct or feature-based methods. The addition of an IMU to VO effectively introduces a fusion pipeline that uses state estimation filters such as the extended Kalman filter (EKF) or particle filters; this is a filtering-based VIO approach, which encompasses most proposed VIO works \cite{gui_review_2015}. In addition to using a filter, the different sampling rates between the IMU and the camera (i.e. the fusion interval) needs to be synchronised, as IMUs generally sample at a rate that is several times faster than the camera's frame rate. This synchronisation is typically achieved by the resampling of the IMU data at the camera's frame rate \cite{konolige_large-scale_2010}. 

OKVIS \cite{leutenegger_keyframe-based_2015}, MSCKF \cite{mourikis_multi-state_2007} and VINS-MONO \cite{qin_vins-mono:_2017} are few of the more popular examples of filtering-based VIO methods in literature; additionally, new VIO methods that are based off existing VO methods such as SVO \cite{faessler_autonomous_2016,forster_-manifold_2017} are often proposed as well. Delmerico and Scaramuzza \cite{delmerico_benchmark_2018} recently published a benchmark comparing monocular applications of these VIO methods across several hardware platforms to measure their performances for autonomous aerial drones. These methods were tested on the EuRoC MAV \cite{burri_euroc_2016} dataset, which consists of visual-inertial sequences that were recorded off an aerial drone. Results from these benchmarks concluded that algorithms which result in higher accuracy and robustness generally require higher computation requirements, thereby implying that VIO methods have to be carefully selected and optimised for their specific applications. VIO methods are often applied to autonomous aerial drones as the addition of an IMU sensor enables estimations up to the six degrees of freedom (6DoF) that is required for their functions. Additionally, a recent method for stereo VIO was presented by Sun et al. \cite{sun_robust_2018}, which postdates this benchmark with the proposal of the stereo multistate constraint Kalman filter (S-MSCKF) method. FAST features are tracked using KLT tracking with RANSAC to remove outliers. Comparisons against OKVIS, ROVIO, VINS-MONO on the EuRoC dataset and on an autonomous aerial drone resulted in the S-MSCKF was able to achieve a balance between computation footprint and accuracy, which suggests that it can be applied on cost-sensitive platforms. While applications on aerial drones are different from autonomous ground vehicles, the benchmarks and works done by \cite{delmerico_benchmark_2018} and \cite{sun_robust_2018} can certainly be used when choosing the proper VIO method for implementation.

A stereo application on mobile ground robots is described by Liu et al. \cite{liu_stereo_2016}, where they have implemented a stereo VIO method onto a model RC car. The VIO method that was proposed utilises multiple Kalman filters for position, orientation and altitude for increased robustness. As the IMU used in the application is considered to be low-cost, the authors also proposed a cascading fusion architecture to estimate orientation measurements, as well as using linear subfilters with low computational footprints with the intention of an embedded computer application. The stereo VO method is a feature-based one whereby SURF descriptors are tracked from the feature pair from each camera, which is detected using the CenSure detector; outlier removal is done using RANSAC. The proposed VIO method was first tested on the KITTI dataset and noted that a pure VO approach will fail at certain turning corners, which is thus rectified with a VIO approach. The second test was performed in a real-world campus pedestrian environment using the mobile robot, where the authors noted that their VIO approach has the least closed-loop error when compared against other state-of-the-art methods, yielding high accuracies while reducing IMU drift errors. While the authors stated that an implementation on an embedded computer with other sensors such as the GPS is part of their future work, the relation to this and autonomous driving is significant, whereby the same method can be adapted for urban road environments, especially when it is benchmarked on the KITTI dataset.

More recently, the use of event cameras for VIO has been proposed by Vidal et al. \cite{vidal_ultimate_2018}. These cameras are capable of high frame rates and are insusceptible to motion blur, as they transmit pixel intensity changes and not the intensity itself. This method achieves VIO with both an event camera and a standard camera, by tracking FAST corners using the KLT tracker, as well as using the Ceres Solver \cite{agarwal_ceres_nodate} for optimisation, which is a non-linear optimizer that selects between the event camera and the standard camera for VO, based on its current environmental conditions around the keyframe. Evaluations were performed on the Event Camera Dataset \cite{mueggler_event-camera_2017} comparing results from using frames (standard camera), events (event camera) and the IMU, whereby the combination of events, frames and IMU yielded significantly better results. Real-world tests on an aerial drone were tested to be resilient against sudden illumination changes, with accurate positioning even in low-light conditions. The high frame rate and robustness achieved using this sensor combination yields results that are real-time and accurate. These results will be highly beneficial to autonomous driving where precise positioning and real-time frame calculations are required. However, the lack of urban road testing by the authors implies that this method should be replicated to assess its capabilities in an autonomous driving environment. 

\section{Discussions}\label{secdiscuss}
This section discusses our overall observations and deductions with regards to the works that were reviewed that are in-line with the current research trends in VO, as well as the requirements for VO to be implemented in autonomous driving applications.  

An application for autonomous driving will require that the implemented methods are capable of running in real-time; in the case of visual algorithms such as VO, 10 Hz is the minimum desired frame rate in order for the vehicle to sustain driving in urban environments \cite{bojarski_end_2016}, while some are explicitly optimised for real-time applications \cite{jaimez_fast_2017,holzmann_detailed_2016}, other methods could be further optimised for real-time performances. It should also be noted that the speed performances of VO algorithms can be further improved just through the use of hardware with higher specifications such as high-performance GPUs. Additionally, in order for a VO method to be accurately robust for autonomous driving, the method should account for the various environmental dynamism that occurs on road scenes. This includes moving obstacles, scene changes and illumination invariances. This was noted in \cite{an_semantic_2017,kim_effective_2016} where dynamic changes in the road scene will affect the accuracies of VO. Through our studies, this dynamic road scene problem can be addressed either through an outlier rejection model such as RANSAC or PASAC \cite{liu_robust_2017}, or through a semantic selection of classified objects \cite{an_semantic_2017}. We observed that while the object classification approach might yield higher accuracies due to its deep learning back-end, using an outlier rejection method is computationally simpler and it is more capable of the real-time performance required for autonomous driving.

We observed an insufficiency in real-time VO methods that are explicitly implemented in an autonomous vehicle. While there exist VO implementations on vehicles such as on the parking camera \cite{lovegrove_accurate_2011}, to the best of our knowledge VO implementations for autonomous vehicles are unavailable. Most autonomous vehicle developments utilise the camera for lane-keeping and obstacle detection/avoidance. However, effective autonomous driving will require precise vehicle localisation and dead-reckoning in the order of centimetres for it to navigate even in unmapped environments. This precision is unattainable through conventional GPS receivers with accuracies of approximately 10 metres; expensive differential GPS receivers are typically installed to achieve this but it introduces redundancy when VO algorithms themselves can be utilised for accurate localisation. Another option to achieve relative localisation is through the installation of an IMU, but unless the IMU is highly precise, the drift errors introduced by an IMU will exponentially affect accuracies; hence the proposals of visual initial odometry. The lack of real-world implementations can also be credited to local legislation and the lack of development vehicles for autonomous drive tests. Here we see that almost all of the recent VO works presented are tested on datasets, and while public datasets such as KITTI or TUM is a great platform and yardstick for method comparisons, it remains to see how these methods will eventually perform in the real world. Testing on these datasets also limit the testing environments to the location where the dataset is captured and is not an effective indicator of the performances of these algorithms in other cities and countries, which will introduce different road scenes altogether. Another reason for the lack of real-world implementation is likely due to the higher computation requirement of VO, which implies that an on-line implementation will require a computer with dedicated and adequate parallel processing hardware. Commercial autonomous driving computers such as Nvidia's DRIVE PX2 \cite{nvidia_corporation_autonomous_nodate} are expensive and are generally unaffordable for developments on a budget, whereas mobile computers such as laptops do not possess the parallel computing capabilities of desktop GPUs. However, the recent availability of low-cost, high performance embedded computers such as the Nvidia Jetson \cite{nvidia_corporation_embedded_2017} and the optimisations of fast VO methods \cite{jaimez_fast_2017,jaimez_fast_2015,steinbrucker_real-time_2011,sun_robust_2018,wu_framework_2017} could catalyse these implementations. 

VO methods are instead often tested on aerial drones and mobile robots, as they usually provide better feasibility and cost-effectiveness as compared to an actual vehicle implementation. VO on an aerial drone is more complex and often fused with inertial measurements as it has more DoF than a ground vehicle; however, VO on a mobile ground robot also differs from an autonomous vehicle whereby it is difficult to replicate the dynamism of road scenes outside of a simulated environment or in the real world. 

As we observed that many VO methods evaluated its performances against visual SLAM (VSLAM) algorithms, this should suggest that VO is similar to VSLAM such that VSLAM uses VO to achieve relative localisation. VO is however different from VSLAM whereby it does not perform loop closures that are necessary for area mapping. Since an autonomous vehicle does not rely on mapping for odometry nor localisation, VSLAM is therefore beyond the scope of our applications. 

\section{Conclusion}\label{secconcl}
The availability of recent works is greatly contributing towards the solution for the visual odometry problem. We have reviewed the various types of visual odometry methods in relation to their applications for autonomous driving. Both monocular and stereoscopic VO are viable approaches for autonomous driving, whereby the hardware will only differ according to its camera setup. The easy attainability of publicly available datasets for autonomous driving such as the well-known KITTI dataset is also a major contributing factor that encourages works in VO. By reviewing recent works pertaining to VO that are not more than three years old, we can confirm that the current VO trend is steering towards a low-cost, high accuracy model that encourages applications on low-powered hardware such as embedded computers. Although the availability of datasets is promoting the proposal of new VO algorithms, we have also observed a current shortage of real-world VO applications, especially in autonomous road vehicles. Many proposed methods stop short of a practical implementation, and only evaluated their algorithms on datasets. We have observed that results from a dataset evaluation often deviates from a complete indication of how the algorithm will perform in a local environment, thereby highlighting our necessity for a practical VO application. We have deduced from our observations that a couple of factors could attribute to this issue. Firstly, the attainability of test beds for autonomous vehicles is often associated with high costs and complex local legislation, as it usually involves the purchase and retrofitting of an actual road vehicle, especially when considering the higher probability of accidents when new algorithms are tested. On the other hand, while embedded computers are now capable of efficient parallel processing, they are still often unable to provide the necessary computing performance required for visual navigation on the test bed, especially when we compare them against workstation-class GPUs that are typically used to evaluate newly proposed algorithms. Nonetheless, forthcoming high-performance mobile computers and an increasing public recognition towards autonomous vehicles will undoubtedly encourage practical applications of visual odometry in the near future.

\bibliographystyle{unsrt}  
\bibliography{main}  

%
%
%
%

\end{document}